\documentclass[num-refs]{wiley-article}

\usepackage{siunitx}
\usepackage{multirow}
\usepackage{subfigure}

\papertype{Original Article}

\title{A micromechanics-based recurrent neural networks model for path-dependent cyclic deformation of short fiber composites}

\author[1]{J. Friemann}
\author[2]{B. Dashtbozorg}
\author[1]{M. Fagerström}
\author[3]{S.M. Mirkhalaf}

\affil[1]{Department of Industrial and Materials Science, Chalmers University of Technology, Gothenburg, Sweden}

\affil[2]{Department of Biomedical Engineering, Eindhoven University of Technology, Eindhoven, the Netherlands}

\affil[3]{Department of Physics, University of Gothenburg, Gothenburg, Sweden}

\corraddress{Department of Physics, University of Gothenburg, Gothenburg, Sweden}

\corremail{mohsen.mirkhalaf@physics.gu.se}

\fundinginfo{Swedish Research Council (VR), Grant/Award Number: 2019-04715; Vinnova}

\begin{document}

\begin{frontmatter}
\maketitle

\begin{abstract}
The macroscopic response of short fiber reinforced composites is dependent on an extensive range of microstructural parameters. Thus, micromechanical modeling of these materials is challenging and in some cases, computationally expensive. This is particularly important when path-dependent plastic behavior is needed to be predicted. A solution to this challenge is to enhance micromechanical solutions with machine learning techniques such as artificial neural networks. In this work, a recurrent deep neural network model is trained to predict the path-dependent elasto-plastic stress response of short fiber reinforced composites, given the microstructural parameters and the strain path. Micromechanical mean-field simulations are conducted to create a data base for training the validating the model. The model gives very accurate predictions in a computationally efficient manner when compared with independent micromechanical simulations.

\keywords{Short fiber composites, Micromechanics, Recurrent neural networks, Deep learnin, Path-dependent plasticity, Cyclic deformation}
\end{abstract}
\end{frontmatter}

\section{Introduction}

Short Fiber Reinforced Composites (SFRCs) are popular materials for engineering applications, partly due to being able to be manufactured through injection molding. The discontinuous nature of the fibers prevents loads from being transferred between themselves and thus, the matrix' role is transferring loads. The high stiffness of the fibers results in the potential of high stresses inside the composite. These high stresses often lead to yielding in the matrix. This interplay between the composite constituents leads to a non-linear elasto-plastic mechanical response of the SFRC. Non-linear analysis of SFRCs is therefore highly relevant for predicting the behavior of structural parts.

A wide variety of microstructural parameters, such as fiber volume fraction and fiber orientation distribution (among others), plays an important role in the mechanical performance of these materials. Thus, different homogenization schemes have been extensively investigated in different micromechanical models. Currently, the most accurate homogenization technique is using computational homogenization performed on realistic Representative Volume Elements (RVEs). Despite accurate predictions, it is not always feasible to use this approach due to some challenges such as difficult RVE generation \citep{MIRKHALAF2019} and expensive simulations \citep{MIRKHALAF2020}. Alternatively, two-step and mean-field methods have been developed and used during the last decades (see more information in e.g. \citep{Doghri2005,MIRKHALAF2021})

More recently, data-driven modeling approaches are being investigated. One such approach is machine learning utilizing Artificial Neural Networks (ANN). An ANN typically consists of a large, layered network of artificial neurons whose individual properties are very simple, but their connected behavior can give rise to very complex phenomena. Neural networks have successfully been utilized for constitutive modeling of specific materials (see e.g. \citep{huang} and \citep{zhang}) and have shown great promises. While the referred works are great proofs of concept, they are developed with only one specific material in mind and do not necessarily generalize to nonlinear micromechanical  modeling. Previous studies that have developed micromechanics-based ANN models for general 3D SFRCs with arbitrary fiber orientations and volume fractions, are limited to the elastic regime, cf. e.g. \citep{Mentges2021}. 

In this study, a micromechanics-based ANN model is developed for path-dependent cyclic elasto-plastic behavior of SFRCs. The main focus is given to the capability of ANN for reproducing general non-linear behavior of these materials, rather than analyzing a specific SFRC and direct comparisons to experiments. Although the same constituent materials are assumed, a variety of orientation distributions, fiber volume fractions and loading-paths are considered in the model. As for all machine learning problems, the issue of properly choosing the training data is central. To generate sufficiently general strain inputs, a continuous random walk-scheme in 6D strain space is proposed. The resulting strain paths are each associated with an orientation tensor describing the fiber distribution, and a fiber volume fraction. For the sampling of uniformly distributed orientation tensors, a simple but effective algorithm utilizing constraints on the tensors' eigenvalues is employed. 

The generated inputs are all given a corresponding goal output by performing  Mean-Field (MF)-simulations using the aforementioned inputs. The results indicate that it is feasible to use ANN to model elasto-plasticity for general 3D analysis of SFRC. The proposed method for generating strain paths shows promise and leads to a training set that is general enough to enable prediction of arbitrary, even highly complex, loading histories. The obtained ANN model is able to accurately reproduce non-linear elasto-plastic stress-strain curves for SFRCs with arbitrary fiber orientation distributions and for a reasonable range of fiber volume fractions.

The remaining of this paper is structured as follows. Section \ref{Micromechanical simulations} describes the constituents of the studied SFRCs and generating the data base. Section \ref{Design and training of artificial neural network} explains the design and training of the ANN model. Results obtained from the ANN model and comparisons to micromechanical simulations are given in Section \ref{Results}. Section \ref{Discussion} gives a discussion on the developed ANN model and obtained results. Finally, Section \ref{Conclusions} gives some concluding remarks.   






\section{Micromechanical simulations}
\label{Micromechanical simulations}
In this section, first, the constituent phases (matrix and reinforcements) of the analyzed composites are explained. Then, the method for generating the required data, for training and validating the artificial neural network model, is explained. 





\subsection{Constituents of the studied SFRCs}
We focus on a specific pair of matrix and reinforcements, namely a polymeric matrix (which behaves similarly as a Polyamide 6.6) and short glass fibers.  
With the same matrix and reinforcements, we will consider different fiber orientation distributions and fiber volume fractions under arbitrary loading conditions. Kammoun et al. \citep{Kammoun2011} proposed describing this material with linear elastic fibers and a matrix obeying $J_2$-plasticity with isotropic linear-exponential hardening. The hardening stress is formulated as
\begin{equation}
\label{eq:microhardeningstress}
    \kappa= -H k +H_{\infty}\left(1-e^{mk}\right),
\end{equation}
where $H$ is the linear hardening modulus, $H_{\infty}$ is referred to as the hardening modulus, $m$ is the hardening exponent, and $(-k)$ is the accumulated plastic strain. The yield function is given by
\begin{equation}
    \label{eq:yieldfunction}
    \Phi(\bm{\sigma},\kappa) = \sigma_{\text{eq}}-(\sigma_{\text{y}}+\kappa)\leq 0,
\end{equation}
where $\sigma_{\text{y}}$ is the yield stress and $\sigma_{\text{eq}}$ is the von Mises equivalent stress defined using the deviatoric stress:
\begin{equation}
    \label{eq:vonMises}
    \sigma_{\text{eq}} = \sqrt{\frac{3}{2}\bm{\sigma}_{\text{dev}}:\bm{\sigma}_{\text{dev}}},\quad \bm{\sigma}_{\text{dev}} = \bm{\sigma}-\frac{1}{3}\text{tr}(\bm{\sigma})\bm{I}.
\end{equation}
Where $\bm{I}$ is the second order identity tensor. 
%
Kammoun et al. \citep{Kammoun2011} presents some of the model parameters of the matrix material in terms of the matrix yield stress $\sigma_{\text{y}}$. The yield stress itself is not given as classified information. In the current work, the value of $\sigma_{\text{y}}=25\ \text{GPa}$ is used. This seems to be a typical value for the yield stress for this type of material \citep{Wu2014}. The used material parameters are given in Table \ref{tab:materialparameters}. 
\begin{table}[h!]
    \centering
    \caption{The material parameters used for modeling the SFRCs.}
 \scalebox{0.95}{   \begin{tabular}{|c|c|c|}
        \hline
        & \textbf{Parameter} & \textbf{Value}\\ \hline \hline
        \multirow{4}{*}{Fiber} &Young's modulus $E_{\text{F}}$ & 76 GPa\\ \cline{2-3}
        &Poisson's ratio $\nu_{\text{F}}$ & 0.22\\ \cline{2-3}
        &Length $l$ & $240\ \mu$m\\ \cline{2-3}
        &Diameter $d$ & $10\ \mu$m\\ \hline \hline
        \multirow{6}{*}{Matrix} &Young's modulus $E_{\text{M}}$ & 3.1 GPa\\ \cline{2-3}
        &Poisson's ratio $\nu_{\text{M}}$& 0.35\\ \cline{2-3}
        &Yield stress $\sigma_{\text{y}}$& 25 MPa\\ \cline{2-3}
        &Linear hardening modulus $H$ & 150 MPa\\ \cline{2-3}
        & Hardening modulus $H_{\infty}$&20 MPa\\ \cline{2-3}
        &Hardening exponent $m$&325\\ \hline
    \end{tabular}  }
    \label{tab:materialparameters}
\end{table}
%
\subsection{Generation of data base}
The generation of the network training data requires three steps, \textbf{(i):} generating strain paths that cover a representative spectrum of loading conditions (described Subsection \ref{Generation of strain paths}); \textbf{(ii):} generating a second order orientation tensor with a corresponding fiber volume fraction for each strain path (described Subsection \ref{Uniform sampling of orientation tensors}); \textbf{(iii):} conducting strain-controlled micromechanical mean-field simulations (described in Subsection \ref{Mean-field simulations}). 
When training the network, the generated strain paths, orientation data, and fiber volume fractions are fed as inputs, and the corresponding stress responses pose as the quantity being predicted. In Subsection \ref{Implementation}, the implementation of the aforementioned steps is explained.

\subsubsection{Generation of strain paths}
\label{Generation of strain paths}
In order to be able to build a network that can learn the proper path-dependency of plasticity, representative strain paths need to be generated. The paths used for training need to be varied enough for the network to be able to recognize complicated paths without losing the ability to predict common simple situations such as uniaxial stress. Randomly sampled data is a useful approach in ensuring a good distribution of data. The developed algorithm is explained in details in what follows. 

\begin{figure}[h!]
    \centering
    \includegraphics[width=0.90\textwidth]{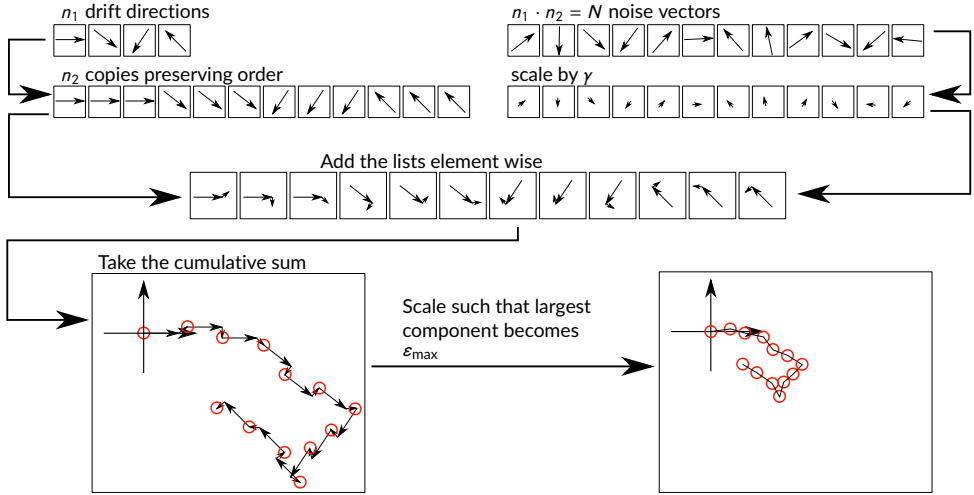}
        \put(-215,68){\small{Scale such that largest}}
        \put(-215,59){\small{component becomes}}
        \put(-215,52){\small{$\varepsilon_{\text{max}}$}}
        \put(-344,180){\small{$n_1$ drift directions}}
        \put(-344,156){\small{$n_2$ copies preserving order}}
        \put(-174,180){\small{$n_1\cdot n_2= N$ noise vectors}}
        \put(-174,156){\small{scale by $\gamma$}}
        \put(-246,123){\small{Add the lists element wise}}
        \put(-330,85){\small{Take the cumulative sum}}
        \caption{Schematic representation of the proposed strain path generation algorithm (for using $n_1 = 4$ and $n_2 = 3$).}
    \label{fig:pathalgo}
\end{figure}

In order to sample 6-dimensional vectors uniformly on a unit 6D-sphere, the following procedure is employed. First, the components of the 6-dimensional vector are sampled independently from each other. The samples are picked from a normal distribution with mean $\mu = 0$ and standard deviation $\sigma^2 = 1$. Thereafter, the vector is normalized such that its magnitude is 1. Since the product of independent identical normal distributions is a normal distribution (which possesses radial symmetry), the resulting probability distribution will be uniform on the unit sphere after the normalization.
Let $N$ be the number of steps in the strain time series (total length $N+1$ including the initial position at the origin). Subsequently, let $n_1$ be the number of drift directions sampled for the random walk and let $n_2$ be the number of steps per drift direction. These parameters are supplied with the constraint $n_1n_2=N$. Also, let $\gamma$ be a number between 0 and 1 that determines the relative amplitude of the perturbations. The algorithm begins with sampling an ordered list of $n_1$ vectors, these vectors are the drift directions. Every vector in the list is repeated $n_2$ times such that the order is preserved, i.e the list with now $n_1\cdot n_2 = N$ elements becomes: $\bm{v}_1,...,\bm{v_1},...,\bm{v}_{n_1},...,\bm{v}_{n_1}$. Following this, $N$ additional vectors are sampled and put into a second ordered list, each one of these new vectors are scaled by the factor $\gamma$. A third list is constructed by taking the element-wise vector sum of the two lists, each containing $N$ elements. A 6-dimensional time series is obtained by taking the cumulative sum of list number three. The time series describing the evolution of the strain components finally follows by re-normalizing such that the largest magnitude element is some desired value $\varepsilon_{\text{max}}$. The presented algorithm (reduced to 2 dimensions for the sake of easy visualization) is displayed in Figure \ref{fig:pathalgo}.

\subsubsection{Uniform sampling of orientation tensors}
\label{Uniform sampling of orientation tensors}
\begin{figure}[h!]
    \centering
    \includegraphics[width=0.8\textwidth]{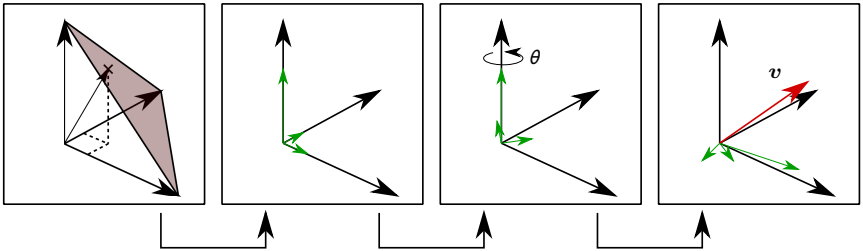}
    \put(-125,70){$\theta$}
    \put(-35,65){$\bm{v}$}
    \caption{The proposed algorithm for uniform sampling of second order orientation tensors.}
    \label{fig:Otalgo}
\end{figure}

The trace of the second order orientation tensor must be unity \citep{Advani1987}, and its eigenvalues must be non-negative. To uniformly sample from the set of all possible orientation tensors, one may uniformly sample triplets of eigenvalues and thereafter perform a coordinate transformation through a randomly chosen rotation. A visualization of the proposed sampling algorithm can be seen in Figure \ref{fig:Otalgo}.

Uniform sampling of three eigenvalues between 0 and 1 that sum to 1 can be achieved by sampling uniformly from the standard 2-simplex. 
This is can be done by sampling 2 points uniformly in (0, 1), and using the length of the three segments these two points make up together with 0 and 1 as the three eigenvalues \citep{devroye1986non-uniform}.

The uniform random sampling of the rotation is performed with a fast and easy to implement algorithm by Arvo \citep{Arvo1992}: First, an angle $\theta$ is uniformly sampled from $(0,2\pi)$, and a single axis rotation $\bm{R}(\theta)$ is performed. Thereafter, a unit vector $\bm{v}=[\cos{\varphi}\sqrt{z},\sin{\varphi}\sqrt{z},\sqrt{1-z}]^{\text{T}}$ is sampled, where $\varphi$ is sampled uniformly from $(0,2\pi)$ and $z$ is sampled uniformly from $(0,1)$. A negatively scaled Householder transformation $-\bm{H}(\bm{v})=2\bm{v}\bm{v}^{\text{T}}-\bm{I}$ is performed, where $\bm{I}$ is the identity matrix. It is possible to show that this transformation will map the axis of rotation to any arbitrary point on the unit sphere. A uniformly sampled rotation $\bm{R}'$ is thus received through the composition of the transformations; $\bm{R}'(\theta,\varphi,z)=-\bm{H}(\varphi,z)\bm{R}(\theta)$.

\subsubsection{Mean-field simulations}
\label{Mean-field simulations}
The stress response pertaining to a given strain path, fiber distribution and fiber volume fraction is computed via micromechanical  simulations. The simulations are performed using a mean-field approach where Mori-Tanaka theory is used for homogenization. The mean-field module of the software \textsc{Digimat-MF} \citep{Digimat} is used to conduct the simulations.
%
%
%
\textsc{Digimat} allows for a first and second order homogenization. 
Second order homogenization shows a significant improvement in accuracy over first order homogenization when there is a large difference between matrix and inclusion stiffness and the matrix exhibits little hardening \citep{Doghri2011}. In the present work these conditions are fulfilled and second order homogenization is therefore chosen. 

\subsubsection{Implementation}
\label{Implementation}
The generation of training data is controlled via a small program written in Julia programming language \citep{julia}. The process of running \textsc{Digimat} simulations is automated by running the program in batch-mode.

The script first generates the strain data and orientation tensors used as input according to the presented algorithms. The total number of steps in the strain paths are set to N = 2000. A pseudo-time variable is also created that starts at 0 s and ends at 1 s with N increments. The time range is completely arbitrary due to the rate independence of the plasticity, and is simply chosen to match the default settings of \textsc{Digimat}. The motivation for long time sequences as training data is that an ANN trained on long sequences would be able to make more accurate predictions compared to shorter input sequences. This follows from the fact that the short sequences are contained in the long ones owing to the construction of the generation algorithm. However, the length of the sequences is a trade-off between generality and computational cost.

For each generated strain path, the generation parameters are sampled randomly. The noise multiplicative factor $\gamma$ is chosen between 0 and 1 from a uniform distribution. The number of drift directions $n_1$ is chosen uniformly from the set $\{1,2,5,10,20,25,50,100,200\}$. This set of number
of drift directions is chosen arbitrarily, motivated only by allowing a fairly large range of path complexities, where $n_1=1$ would mean essentially a perturbed linear function and $n_1=200$ would result in a highly complex path. Furthermore, the maximum admissible strain component $\varepsilon_{\text{max}}$ is sampled uniformly between 0.01 and 0.05. This range captures a full range of reasonable loading conditions. The lower bound allows loading that does not lead to plasticity and the upper limit of 0.05 is kept since failure is extremely likely to occur beyond that point which is outside the scope of this work. As a final parameter regarding the strain paths, the possibility of uniaxial strains is implemented. This is realized by setting all but one of the strain components of a strain path to constant 0 with a probability of 10\%.

For every strain path, a corresponding orientation tensor is generated according to the proposed algorithm. The case of uniaxial fiber alignment is chosen with a probability of 10\% during generation, i.e. one eigenvalue is set to exactly 1 while the remaining two are set to exactly 0. As an additional parameter associated with the fiber distribution, the fiber volume fraction $v_{\text{F}}$ is chosen randomly between 10 and 15\%. The volume fraction is allowed to vary since it will be dependent on the position after injection molding. The range of fiber volume fraction is kept small to make the training process of the network easier. All parameters of the strain path and orientation tensor generation are displayed in Table \ref{tab:straingenerationparams}.
\begin{table}[h!]
    \centering
    \caption{The parameters pertaining to the generation of strain paths and orientation tensors.}
    \scalebox{0.95}{  \begin{tabular}{|c|c|}
         \hline
         \textbf{Parameter}&\textbf{Value}\\\hline \hline
          $N$ & 2000\\ \hline
          $\gamma$ & Uniform random $(0,1)$ \\ \hline
          $n_1$ & Uniformly from $\{1, 2, 5, 10, 20, 25, 50, 100, 200\}$\\ \hline
          $\varepsilon_{\text{max}}$ &Uniform random $(0,0.05)$\\ \hline
          $\mathcal{P}$(uniaxial strain) & 0.1\\ \hline
          $\mathcal{P}$(uniaxial fibers) & 0.1\\ \hline
          $v_{\text{F}}$ & Uniform random (0.1,0.15)\\ \hline
    \end{tabular}  }
    \label{tab:straingenerationparams}
\end{table}


A data set with 40,000 samples is generated. Every sample consists of one strain path, one orientation tensor, one fiber volume fraction, and the resulting stress path. The computations are carried out on a personal computer with an 8 core 3.8 GHz processor and 16 GB of memory. The entire generation process takes about two weeks. The vast majority of the computational time is dedicated to the micromechanical  simulations that compute the stresses. 

\section{Design and training of the artificial neural network}
\label{Design and training of artificial neural network}
The computational programming language \textsc{MATLAB} (MathWorks Inc, Natick, MA) \citep{MATLAB:2020b} is used for the implementation of the artificial neural network. 

To capture the path-dependency of plasticity, Recurrent Neural Networks (RNN) are utilized. 
\textcolor{black}{RNNs are a class of neural networks that allow previous outputs to be used as inputs while having hidden states which makes them powerful for modeling sequence data such as time series.
An improved version of standard RNN,} Gated Recurrent Units (GRU) \citep{cho-etal-2014-learning}, are used as the main neuron type in the network. 

The input signals of the network consist of time series of length $N_T=2001$, with $F=13$ different features. The features comprise the 6 independent components of the second order orientation tensor, the fiber volume fraction, and the 6 independent components of the strain tensor. The time-independent features are simply fed to the network at every time step without modification, while the strain tensor components are changing with time. The output signals are the 6 independent components of the stress tensor. The loss is computed by comparing the output to the stress time series obtained from the micromechanical  simulation corresponding to the input signal. We used the ADAM optimizer \citep{ADAM}, a first order gradient based optimizer with a learning rate of 0.0005.

\begin{figure}[h!]
    \centering
    \includegraphics[width=0.62\textwidth]{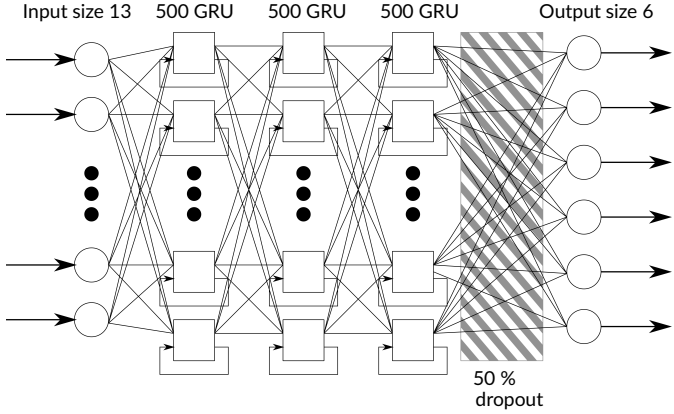}
    \put(-245,135){\small{Input size 13}}
    \put(-195,135){\small{500 GRU}}
    \put(-152.5,135){\small{500 GRU}}
    \put(-110,135){\small{500 GRU}}
    \put(-75,-4){\small{50 \%}}
    \put(-74,-12){\small{dropout}}
    \put(-50,135){\small{Output size 6}}
    \caption{The proposed network architecture.}
    \label{fig:networkarchitecture}
\end{figure}

As illustrated in Figure \ref{fig:networkarchitecture}, the network structure begins with an input layer of time sequence data. After the input layer, three layers follow each including 500 GRU neurons. In order to give the signal the correct output dimension, and to match the magnitude of the output training data, a fully connected feed-forward layer is placed after the GRU layers. This layer has unit activation such that it can reproduce stresses of arbitrary magnitude and sign. For network regularization and over-fitting prevention, a dropout layer \citep{srivastava14a} is added between the final GRU layer and the feed-forward layer (represented by the dashed box) with a dropout probability of 50\%. The final layer is a fully connected layer with 6 neurons that emits the output time series. 

The 40,000 generated data samples are randomly split into training (80\%), validation (19.75\%) and test (0.25\%) sets. 
\textcolor{black}{The entire training dataset is passed to the same neural network multiple times in an iterative manner (Epochs). At every epoch, the training data is shuffled to ensure that the training process is independent of the structure of the input data.}
%
\textcolor{black}{The validation set is used to evaluate the network performance, and to tune the hyper-parameters such as batch size, learning rate and regularization parameters to control the behavior of the model and to achieve the highest performance.}
The validation set loss at each epoch indicates if the model generalizes its prediction capability to data outside the training set. A validation cost that stops decreasing, or starts to increase, is an indication of network over-fitting.
The batch size is a very important hyper-parameter that influences the convergence rate substantially. Simply taking as large mini-batch size as possible does not necessarily guarantee convergence to the global minimum \citep{keskar2017largebatch}. During hyper-parameters tuning, a batch size of $N_{\text{batch}}=32$ is chosen.

In order to improve the prospects of convergence, piece-wise learning rate decay is used. The scheduling is motivated by the following: By setting the learning rate too low from the beginning, convergence may be too slow, or the optimizer gets stuck in a local minimum early. On the other hand, setting a high learning rate might help the optimizer arrive in the vicinity of the global minimum. However, the large step size may prevent the optimizer from making the final stretch without overstepping the minimum, that may be in a small dip in an otherwise flat area of the cost function. By scheduling the learning rate one may get the best of both worlds. The learning rate is scheduled to decrease by 10\% every 10 epochs. 

Convergence is usually faster if the average of the input variables is close to zero and that the input variance is close to one \citep{LeCun2012}. The input features are therefore subjected to $z$-score normalization. 
%
In order to prevent over-fitting we used $L_2$-regularization \citep{murphy2012machine}, which modifies the cost function such that it penalizes weights of large magnitude. The penalty term is the squared sum of all the weights multiplied by a factor $\lambda$ that controls the amount of regularization. The value of $\lambda = 0.0001$ is used in the training of the current network. The idea is that weights with large magnitude make the model more complex and reduces its ability to generalize and is therefore penalized. In order to counteract exploding gradients that lead to numerical issues, Gradient clipping is used, which re-scales a gradient if its norm is larger than a set threshold value. 

For the cost function for the sequence regression, we used mean squared error:
\begin{equation}
    \label{eq:matlabcost}
    C = \frac{1}{N_{\text{batch}}}\sum_{n=1}^{N_{\text{batch}}}C_n,\quad \text{where}\quad C_n = \frac{1}{2N_T}\sum_{i=1}^{F} \sum_{j=1}^{N_T}  (\hat{y}^{n}_{ij}-y^{n}_{ij})^2.
\end{equation}
Here $\hat{y}^{n}_{ij}$ and $y^{n}_{ij}$ are the $n$th prediction and target values in a training batch respectively.

The test set, as unseen data by the network, is used to evaluate the network performance after the training has been finished.  The test set is intentionally kept very small in this work since it is only to be used for final validation before proper model testing is carried out. The random nature of the generated paths is not representative of the model use cases. Instead, the model is rigorously tested on more conventional data specifically generated for testing (see Subsection \ref{One cycle tests}). 

\section{Results}
\label{Results}
This section begins with introducing a physically relevant error metric for evaluating the performance of the developed ANN model. Following this, the results of applying the trained model on the test set are presented. Thereafter, the obtained model is tested by subjecting a few virtual samples to some representative load cases. Micromechanical simulations are performed in \textsc{Digimat-MF} and the output is compared to the ANN model prediction. The goal is to demonstrate that the proposed ANN model can reproduce mean-field predictions with a good accuracy. The required rate independence of the model is also demonstrated. It is also shown that the ANN model is able to make accurate predictions outside the parameter range which was used for training.

\subsection{Evaluation metrics}
Simply referring to the cost defined in Equation \eqref{eq:matlabcost} as an error metric is not sufficient, since a numerically small cost does not necessarily imply a small error if the scale of the output is also small. To give the error a physical significance, new quantities need to be introduced. To remove the dimensionality, the root mean square error is divided by the yield stress of the matrix ($\sigma_{\text{y}}$). This metric is from now on referred to as the Mean Relative Error (MeRE). This metric gives an overall estimate of the model accuracy, it does however not account for large localized errors. Hence, the Maximum Relative Error (MaRE) is also introduced. The MaRE is defined by the maximum absolute error divided by the matrix yield stress. The two error metrics of a time series of length $N_T$ are computed for each stress component through the following equations:
\begin{equation}
    \label{eq:errormetrics}
    \text{MeRE} = \frac{\sqrt{\frac{1}{N_T}\sum_{t=1}^{N_T}(\sigma_{t}-\hat{\sigma}_t)^2}}{\sigma_{\text{y}}} \quad\text{and}\quad \text{MaRE} = \frac{\underset{t}{\max}|\sigma_{t}-\hat{\sigma}_t|}{\sigma_{\text{y}}},
\end{equation}
where $\hat{\sigma}_t$ is a predicted stress component at step $t$ and $\sigma_t$ is the desired stress component at step $t$.

\subsection{Obtained model}
The network was trained during 500 epochs on an Nvidia V100 GPU with 32 GB of VRAM. The training took just above 81 hours using 32,000 training samples.

The average of the MeRE and MaRE is computed over the 100 time series for the test set. The time to make a prediction of a stress time series was always less than 1 second, typically taking about 0.1 s. In comparison, the time to predict the stress response corresponding to the strain signals in the test set using \textsc{digimat-mf} was typically around a minute. The average errors for all 6 stress components are displayed in Table \ref{tab:testseterror}. The small errors indicate the good performance of the network in prediction.
\begin{table}[h!]
    \centering
    \caption{The average MeRE and MaRE computed over the test set.}
    \label{tab:testseterror}
   \scalebox{0.95}{  \begin{tabular}{|c||c|c|c|c|c|c|}
    \hline
        &$\sigma_{11}$ & $\sigma_{22}$ & $\sigma_{33}$ & $\sigma_{12}$ & $\sigma_{23}$ & $\sigma_{13}$ \\ \hline \hline
        \textbf{MeRE} & 0.0582 & 0.0528 & 0.0487 & 0.0469 & 0.0378 & 0.0431\\ \hline
        \textbf{MaRE} & 0.1255 & 0.1137 & 0.1121 & 0.1020 & 0.0874 & 0.0980\\ \hline
    \end{tabular}  }
\end{table}

\noindent \textcolor{black}{\textbf{Remark:} It should be emphasized that the matrix yield stress (25 MPa), which is much lower than the maximum composite stresses obtained in the simulations, is used for defining the error metrics.}

\subsection{Virtual sample testing}
Five orientation tensors together with a corresponding fiber volume fraction were sampled uniformly according to the previously proposed rules. Additionally, samples with a uniaxial fiber distribution (1D), a random planar fiber distribution (2D), and a spatially uniform fiber distribution (3D) were created. These three samples all had the fiber volume fraction of 12\%. The corresponding data related to the samples is presented in Table \ref{tab:samples}
\begin{table}[h!]
    \centering
    \caption{The second order orientation tensors and fiber volume fractions for the virtual samples.}
    \scalebox{0.95}{  \begin{tabular}{|c||c|c|c|c|c|c|c|c|c|c|}
    \hline
        \textbf{Sample} & $a_{11}$ & $a_{22}$ & $a_{33}$ & $a_{12}$ & $a_{13}$ & $a_{23}$ & $v_{\text{F}}$  \\ \hline \hline
        1 & 0.477 & 0.188 & 0.335 & -0.080 & -0.071 &-0.183 & 0.130\\ \hline
        2 & 0.094 & 0.692 & 0.214 & -0.103 & 0.012 & -0.255 & 0.144\\ \hline
        3 & 0.649 & 0.139 & 0.212 & 0.011 & -0.117 & -0.154 & 0.131\\ \hline
        4 & 0.392 & 0.225 & 0.382 & -0.142 & 0.080 & 0.152 & 0.139\\ \hline
        5 & 0.000 & 0.919 & 0.081 & 0.015 & 0.005 & 0.273 & 0.109\\ \hline
        1D & 1.00 & 0.000 & 0.000 & 0.000 & 0.000 & 0.000 & 0.120\\ \hline
        2D & 0.500 & 0.500 & 0.000 & 0.000 & 0.000 & 0.000 & 0.120\\ \hline
        3D & 0.333 & 0.333 & 0.333 & 0.000 & 0.000 & 0.000 & 0.120\\ \hline
    \end{tabular}  }
    \label{tab:samples}
\end{table}

The stress-strain responses of the samples for some representative stress states were obtained through strain controlled simulations in \textsc{Digimat-MF}. A load cycle is defined by changing the control strains piece-wise monotonically from 0 to $\varepsilon_{\text{c}}$ to $-\varepsilon_{\text{c}}$ to 0, where $\varepsilon_{\text{c}}$ is the maximum control strain. Only the control strains were strictly imposed, \textsc{Digimat-MF} computed the remaining strain components such that a desired stress state was achieved. The micromechanical  simulation acted as a ground truth, and the model error was computed by comparison with the ANN model output. The model used the strains in the simulations as input, together with the respective parameters corresponding to the different virtual samples.

\subsubsection{One cycle tests}
\label{One cycle tests}
Tests consisting of one load cycle were performed on samples 1-5. Five different stress states were investigated, (i): uniaxial stress in the $\sigma_{11}$-direction controlled by imposing $\varepsilon_{11}$, (ii): pure shear in the $\sigma_{12}$-direction controlled by imposing $\varepsilon_{12}$, (iii): a bi-axial stress state in the $\sigma_{11}$-$\sigma_{22}$-directions controlled by imposing $\varepsilon_{11}$ and $\varepsilon_{22}$, (iv): a bi-axial stress state in the $\sigma_{11}$-$\sigma_{23}$-directions controlled by imposing $\varepsilon_{11}$ and $\varepsilon_{23}$, and finally (v): a plane strain state in the $\varepsilon_{11}$-$\varepsilon_{22}$-plane. The maximum control strain was $\varepsilon_{\text{c}}=0.035$ for these tests. The number of steps was simply matched with the output of the micromechanical  simulations. However, to investigate the rate-independence, the proposed tests were repeated with varying degree of linear interpolation/extrapolation in time on the stress-strain data. 

\begin{figure}[h!]
    \centering
    \subfigure[]{\includegraphics[width=0.46\textwidth]{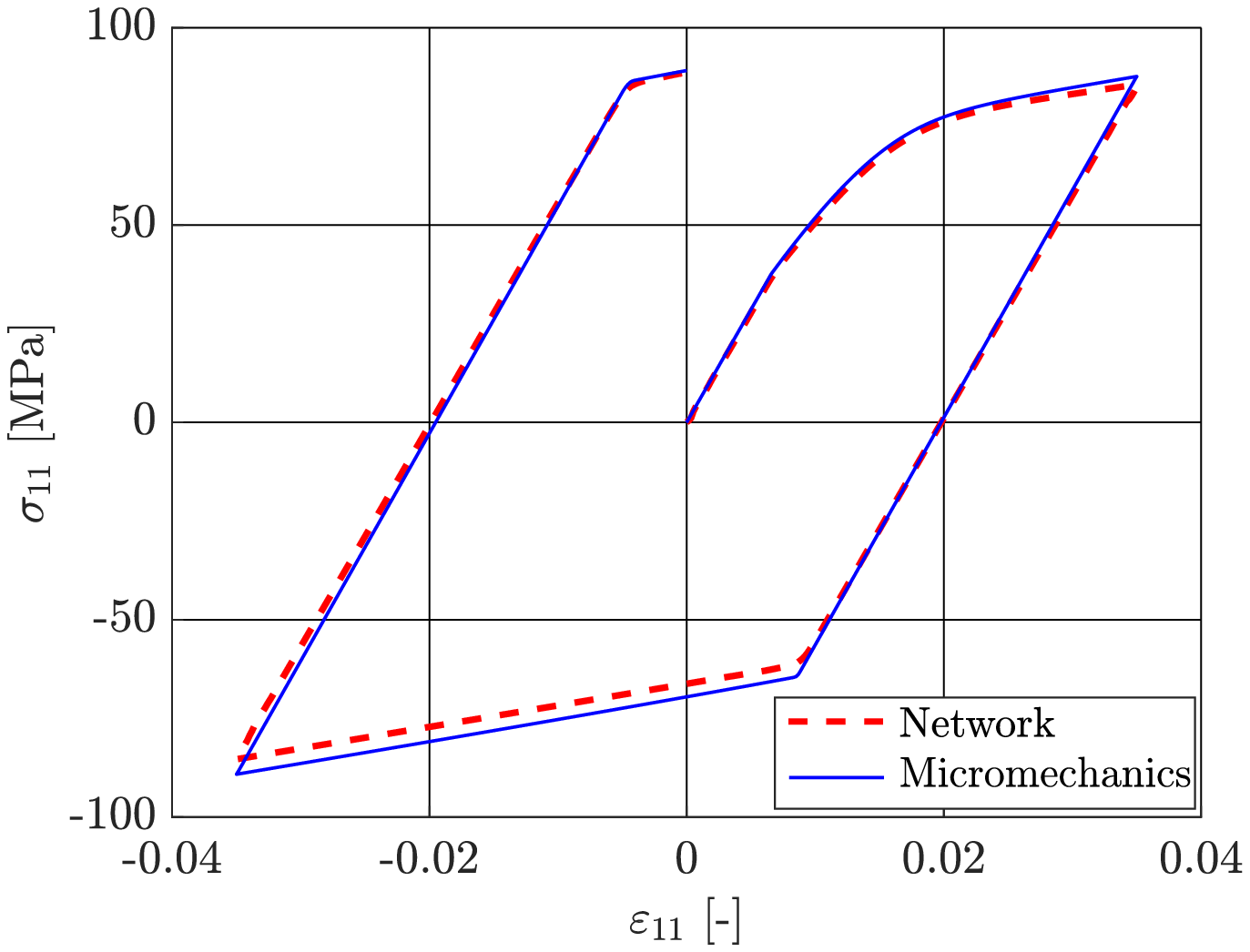}}
     \subfigure[]{\includegraphics[width=0.46\textwidth]{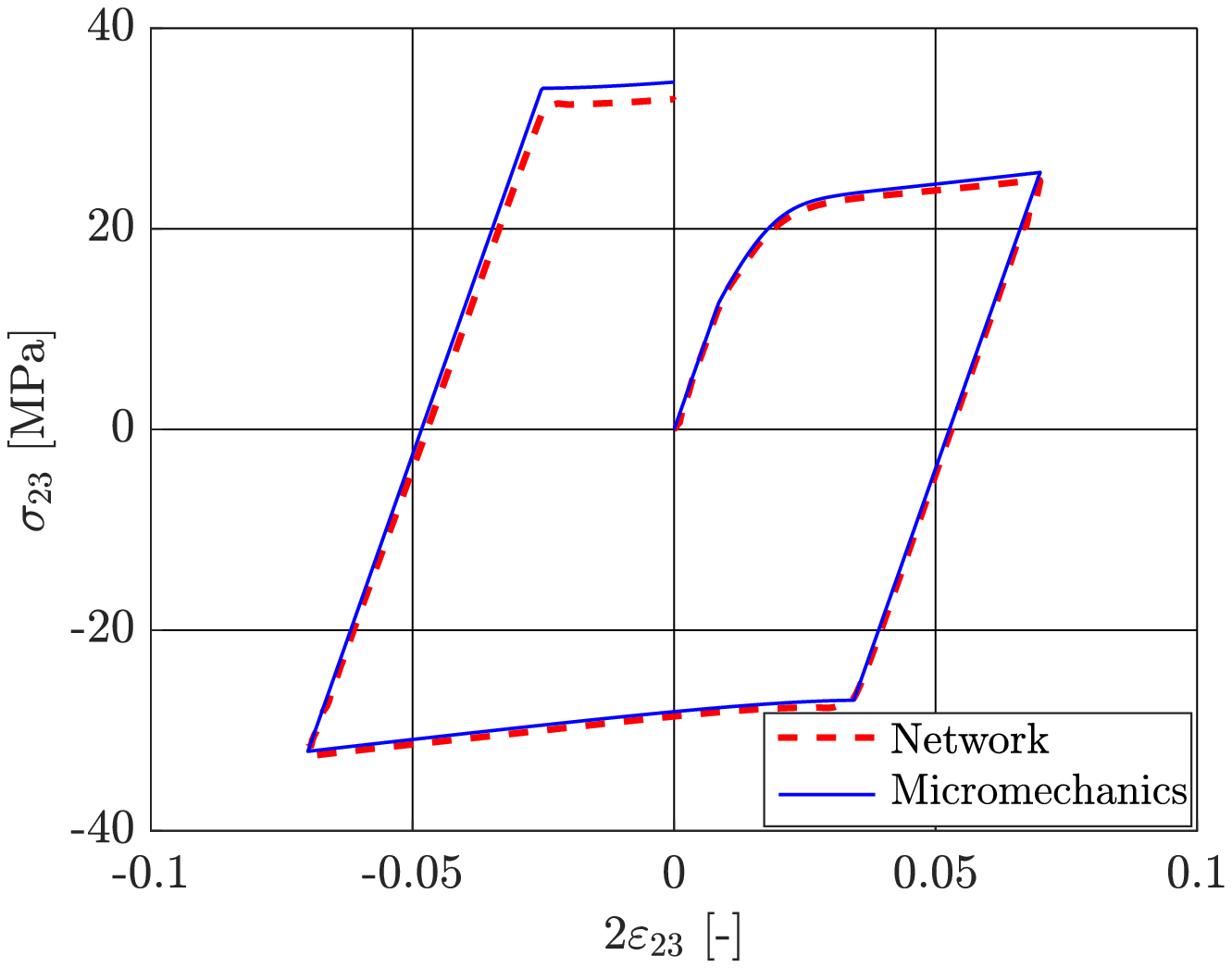}}
        \caption{Comparison of the stress-strain predictions by the network with micromechanics simulations for (a): uniaxial test performed on sample 1, (b): $\sigma_{11}$-$\sigma_{23}$ biaxial test performed on sample 5.}  
        \label{fig:exampleStressStrain}
\end{figure}

The resulting stress-strain curves from the tests show very good agreements between the network predictions and micromechanical simulations. As some representative results, two stress-strain curves for (a) uniaxial test performed on sample 1, and (b) biaxial test performed on sample 5 are shown in Figure \ref{fig:exampleStressStrain}.
These curves highlight that the developed ANN model captures well the fundamental characteristics of the underlying model. Specifically, the model correctly displays a hardening process that saturates for large plastics strains, and the expected loading/unloading behavior.

\begin{figure}[h!]
    \centering
    \includegraphics[width=0.92\textwidth]{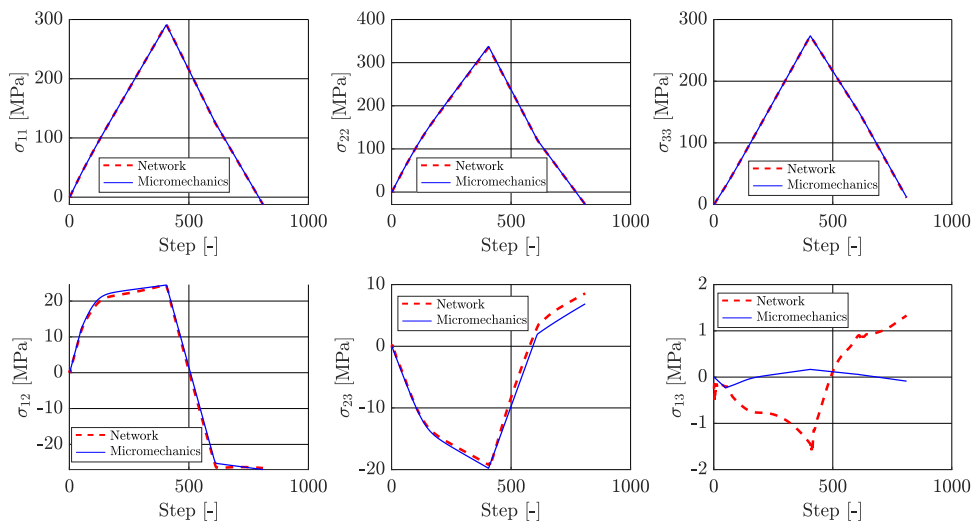}
    \caption{Comparison of the network predictions with micromechanics simulations for the time series data of all 6 stress components for a plane strain state load cycle on sample 5.}    
    \label{fig:sample1_plane_stepStress}
\end{figure}

A demonstrative stress time series of all six stress components for a plane strain test is presented in Figure \ref{fig:sample1_plane_stepStress}. This figure also shows good agreement between the network predictions and the micromechanical simulations. The prediction of the $\sigma_{13}$ looks poor, however by noting the scale of the stress it is apparent that the relative error is still small.
The model shows great promise with a MeRE that never surpasses 25\% of the matrix yield stress during the tests. The component wise average MeRE is also low, less than 7\% of the matrix yield stress. The largest recorded MaRE was less than 50\% of the matrix yield stress, while the component wise average was less than 15\%.

\begin{figure}[h!]
    \centering
    \includegraphics[width=0.92\textwidth]{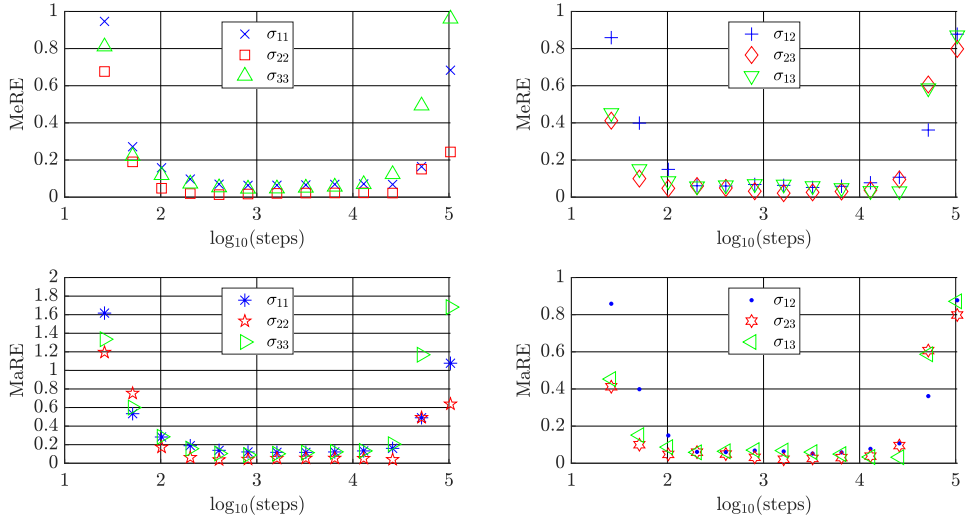}
    \caption{Dependency of the error to sequence length (for a plane strain load cycle on sample 1).}    
    \label{fig:sample1_plane_error}
\end{figure}

Figure \ref{fig:sample1_plane_error} shows the typical behavior of the MeRE and MaRE when the number of steps in the time series is varied. 
It is seen that the errors stay almost constant in the range between 200 to 20,000 steps. But it is quickly growing for sequences shorter or longer than these bounds. This holds true for most of the tests, but some tests showed stricter bounds for a constant error. None of the performed tests showed errors that varied to a significant degree when loaded with sequences of lengths between 500 and 8000 steps.

The limits on the sequence lengths can be explained by the structure of the training data. As the algorithm that was used to generate the training data has a theoretical maximum strain rate, and an average strain rate that is considerably lower than the maximum, it's reasonable that there is a minimum sequence length for which the neural network model can make accurate predictions. This explains the lower bound. Since the sequence length of the training data is fixed, the network is not trained to remember the accumulated plastic strain forever. The loss of accuracy for very long sequences can possibly be explained by the network forgetting about plastic strain accumulated early in the sequence when it approaches the sequence's end. Bearing this in mind, it may still be said that the network model displays rate-independence for a large range of sequence lengths.

\subsubsection{Repeated cyclical loading}
\label{Repeated cyclical loading}
\begin{figure}[h!]
    \centering
    \includegraphics[width=0.92\textwidth]{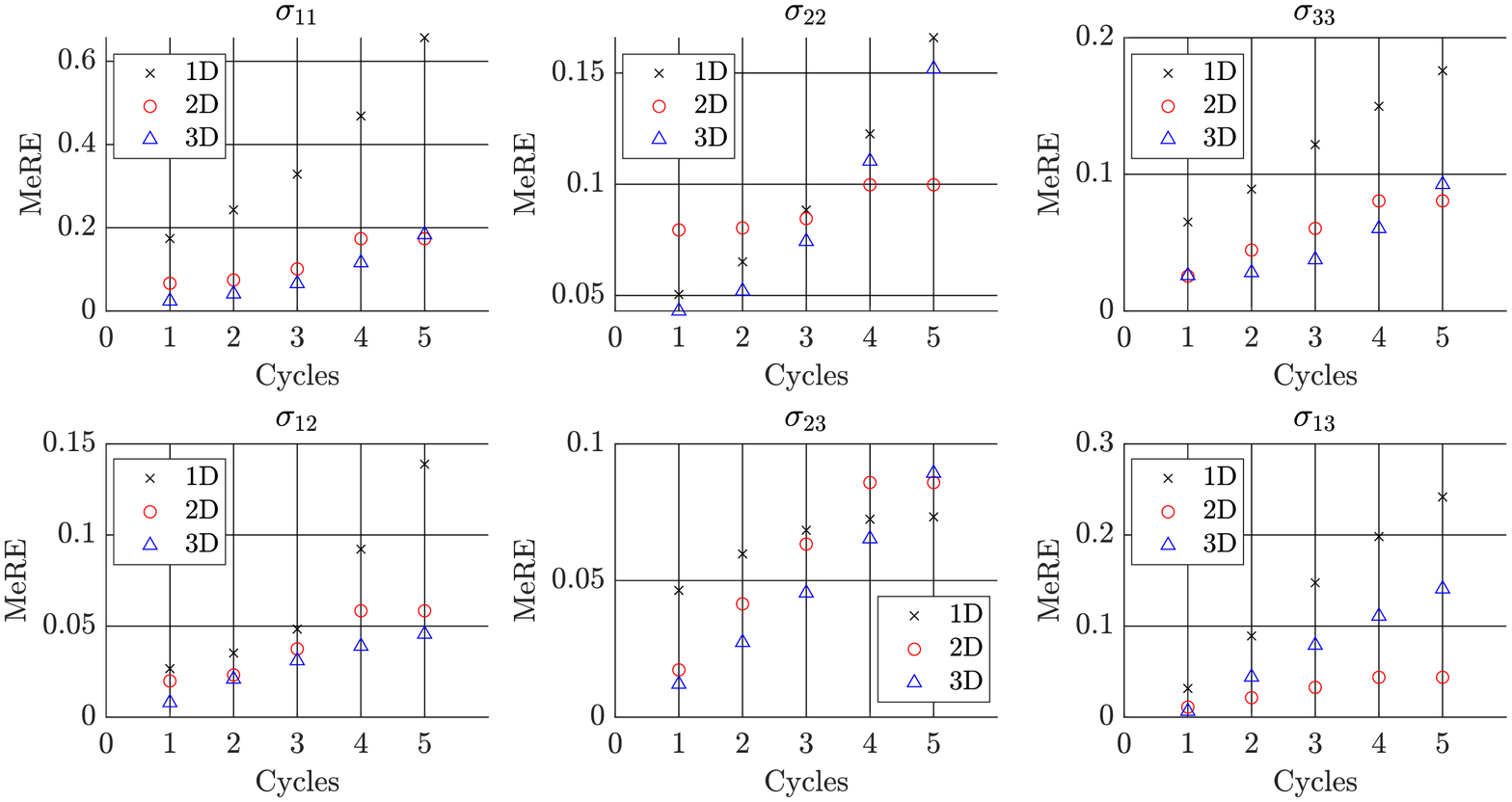}
    \vspace{-5mm}
    \caption{MeRE of cyclical tests on unidirectional (1D), random planar (2D), and spatially uniform (3D) samples as a function of the number of cycles.}    
    \label{fig:cycleMeRE}
\end{figure}

\begin{figure}[h!]
    \centering
    \includegraphics[width=0.92\textwidth]{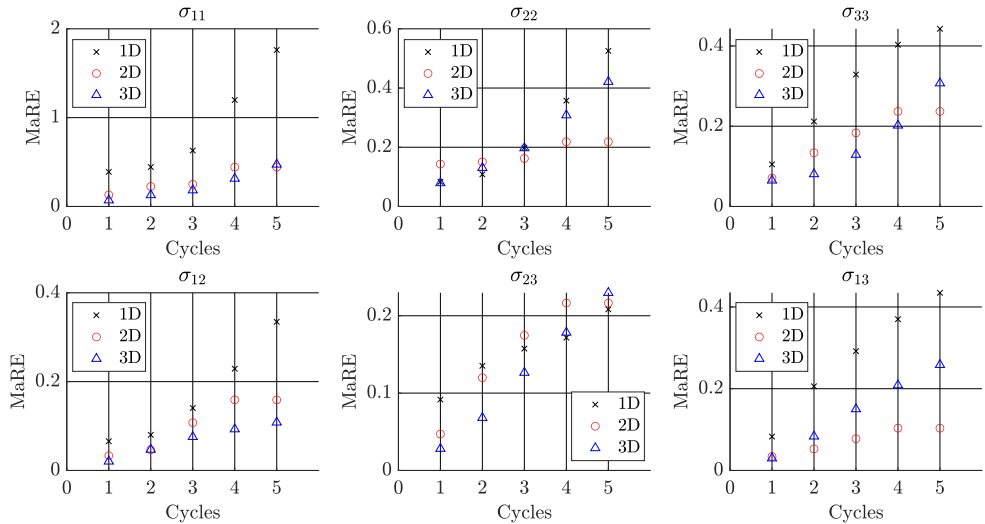}
    \caption{MaRE of cyclical tests on unidirectional (1D), random planar (2D), and spatially uniform (3D) samples as a function of the number of cycles.}
    \label{fig:cycleMaRE}
\end{figure}

To test how complex the load histories can be without forgoing too much accuracy, a test with an increasing number of load cycles was performed. Samples 1D, 2D, and 3D were subjected to a uniaxial stress state in the $\sigma_{11}$-direction. The maximum control strain was $\varepsilon_{\text{c}}=0.04$, and tests of 1 to 5 cycles were performed.
The calculated MeRE corresponding to each of the six strain components as a function of the number of load cycles is displayed in Figure \ref{fig:cycleMeRE}, while the MaRE is found in Figure \ref{fig:cycleMaRE}. The component-wise errors for a uniaxial fiber distribution, a uniform 2D distribution, and a uniform 3D distributions are given.
The error is found to increase almost linearly with the number of load cycles. It can be seen that the uniaxial fiber distribution typically suffers from the greatest error. 
The sequence length is proportional to the number of cycles, where one cycle consists of approximately 800 steps. Since it was shown in the previous section that the error of sequences between approximately 500 and 8000 steps is not affected by changing the sequence length, it can be concluded that the error stems from the complexity of the load path. 

\begin{figure}[h!]
    \centering
    \subfigure[]{\includegraphics[width=0.46\textwidth]{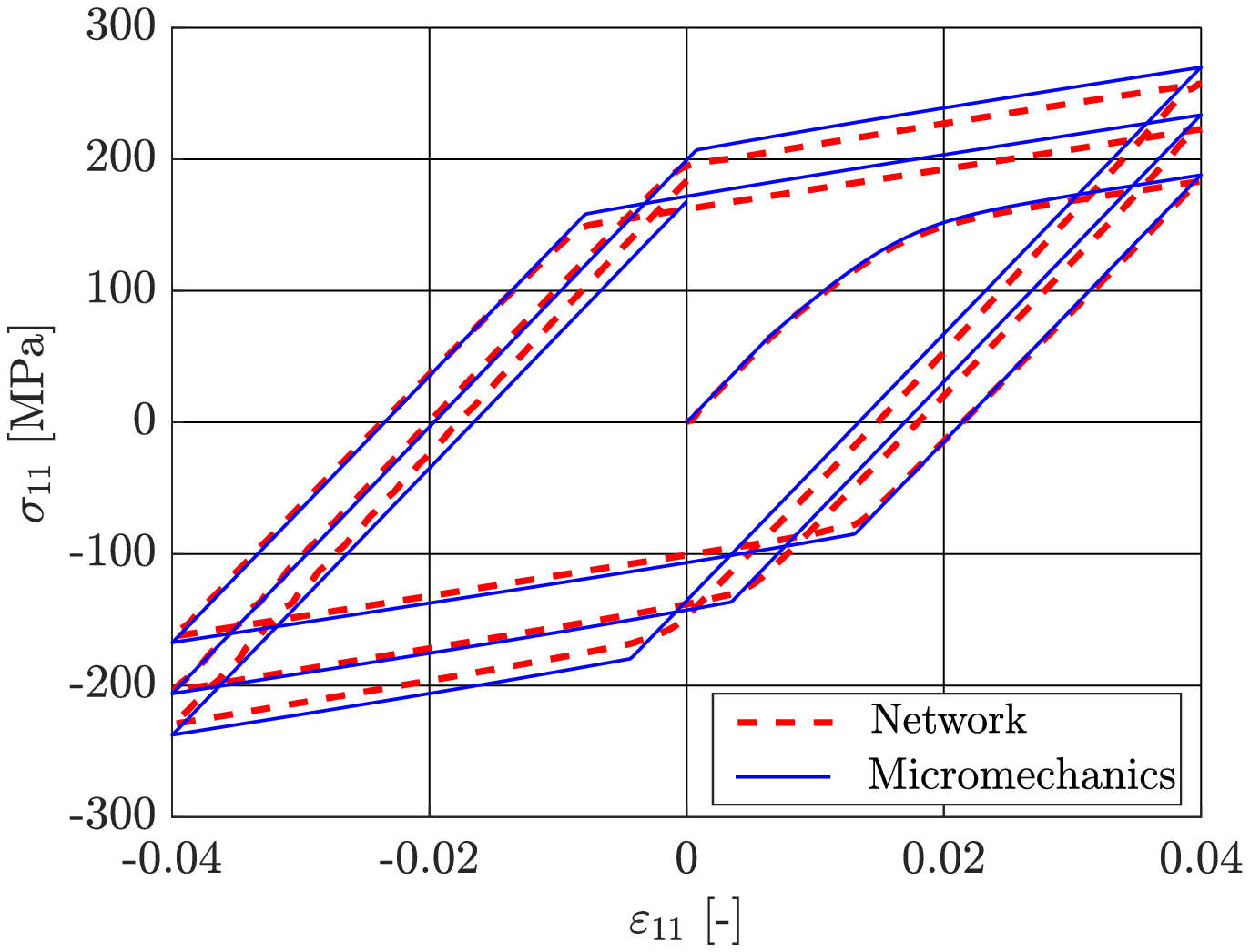}}
     \subfigure[]{\includegraphics[width=0.46\textwidth]{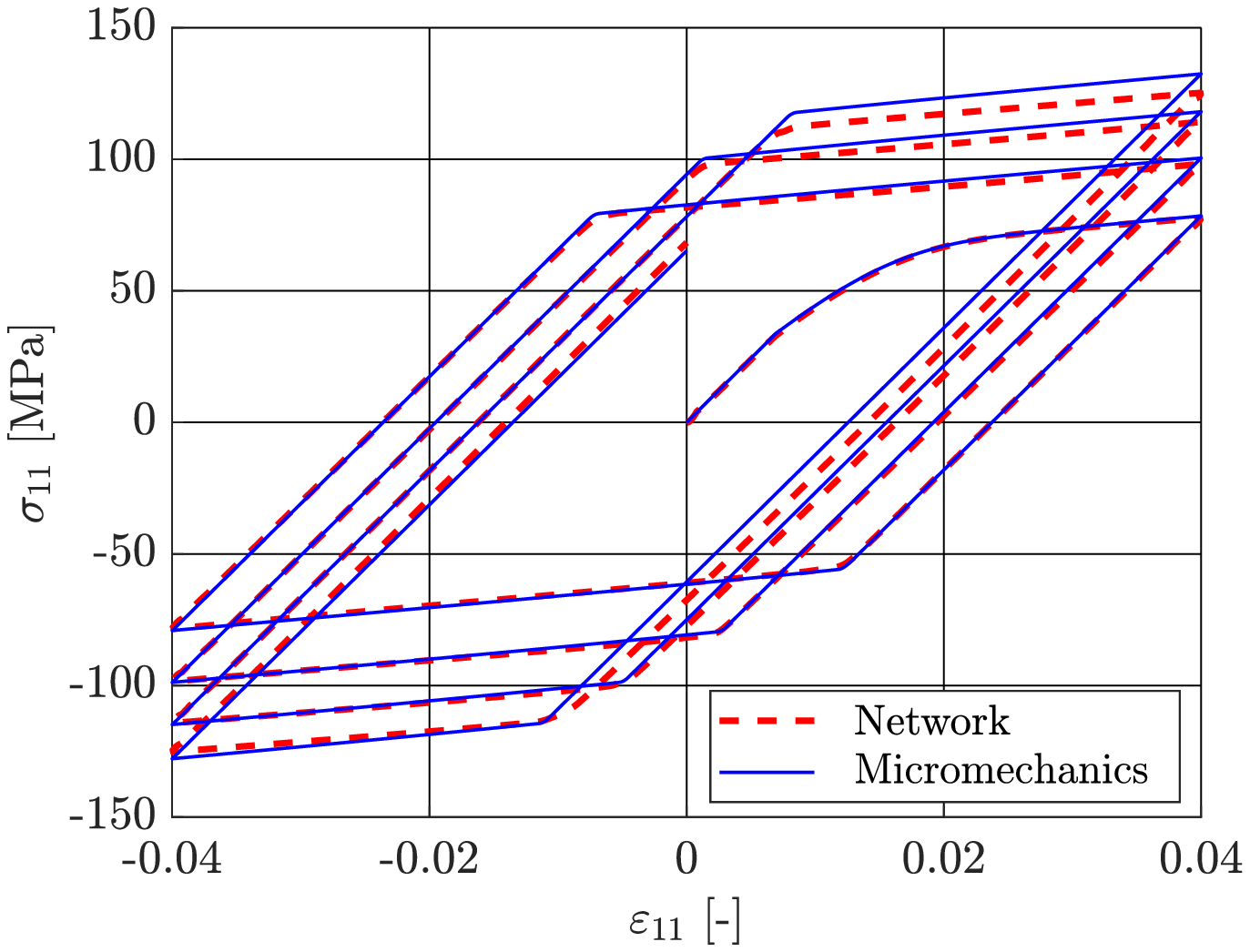}}
        \caption{Comparisons of the stress-strain curves obtained by the network and micromechanics simulations for uniaxial tests performed on samples with (a): a unidirectional fiber distribution, and (b) a uniform 3D fiber distribution.}        
        \label{fig:cycleStressStrain}
\end{figure}

Two representative stress-strain curves from the conducted simulations are displayed in Figure \ref{fig:cycleStressStrain}.
%
It is possible to see how the prediction error increases slightly for every additional cycle. It is also interesting to see that once yielding has occurred, the error seems to stay fairly constant. This speaks in favor of the model accurately capturing material properties such as the tangent modulus.

\subsection{Extrapolation ability}
An additional set of tests consisting of a uniaxial load cycle was performed on a sample with a random 3D fiber distribution. Fiber volume fractions of 0.1\% 2.5\%, 5\%, 7.5\%, 10\%, 12.5\%, 15\%, 17.5\%, and 20\% were investigated. For each volume fraction a maximum control strain ($\varepsilon_{\text{c}}$) of 5\%, 7.5\%, and 10\% was investigated.

\begin{figure}[h!]
    \centering
    \includegraphics[width=1.00\textwidth]{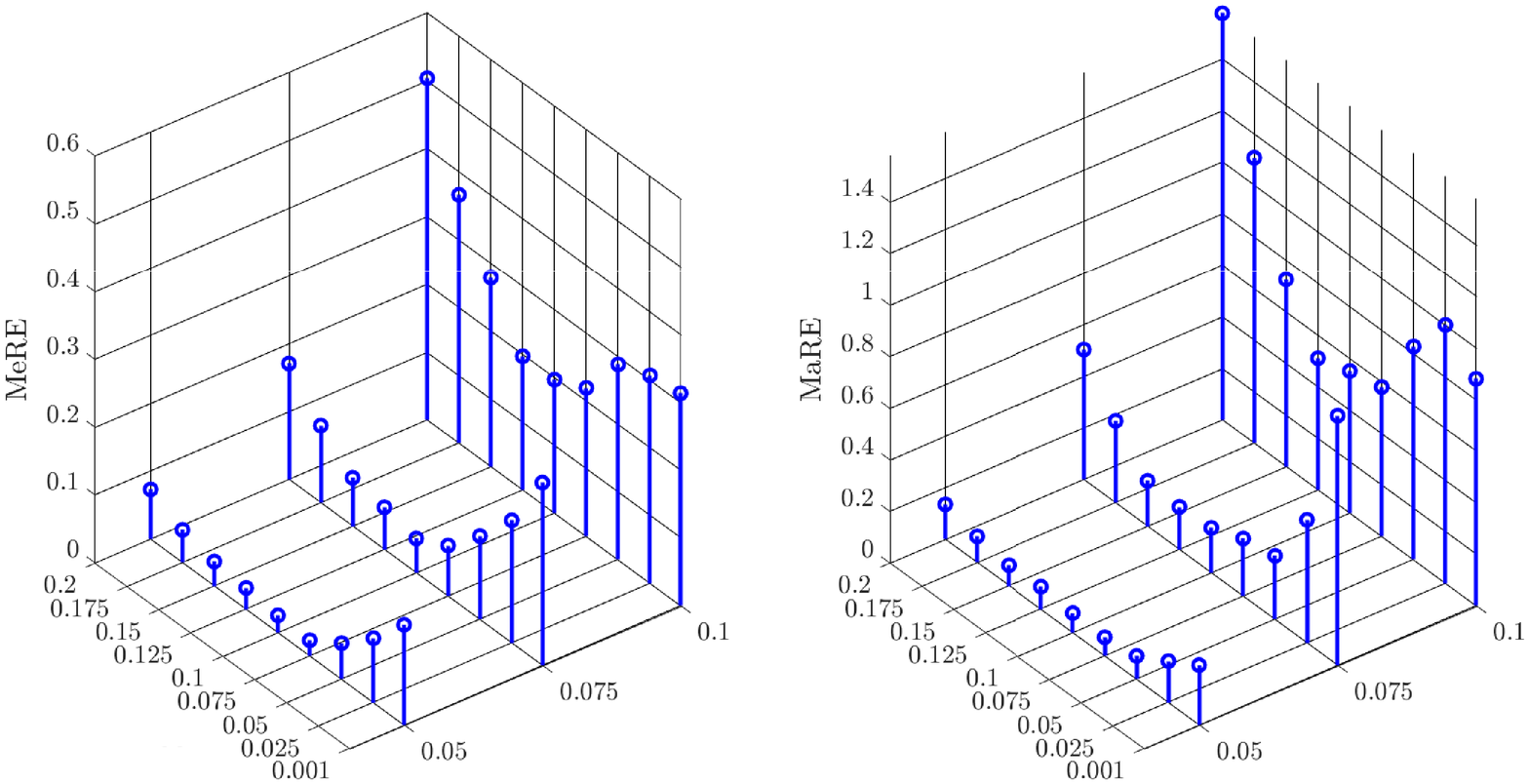}
    \put(-255,20){$\varepsilon_{\text{c}}\ [-]$}
    \put(-70,20){$\varepsilon_{\text{c}}\ [-]$}
    \put(-360,20){$v_{\text{F}}\ [-]$}
    \put(-185,20){$v_{\text{F}}\ [-]$}
    \caption{The error (MeRE and MaRE) in the $\sigma_{11}$ component as a function of maximum control strain and fiber volume fraction.}
    \label{fig:extrapolateError}
\end{figure}

\begin{figure}[h!]
    \centering
    \subfigure[]{\includegraphics[width=0.46\textwidth]{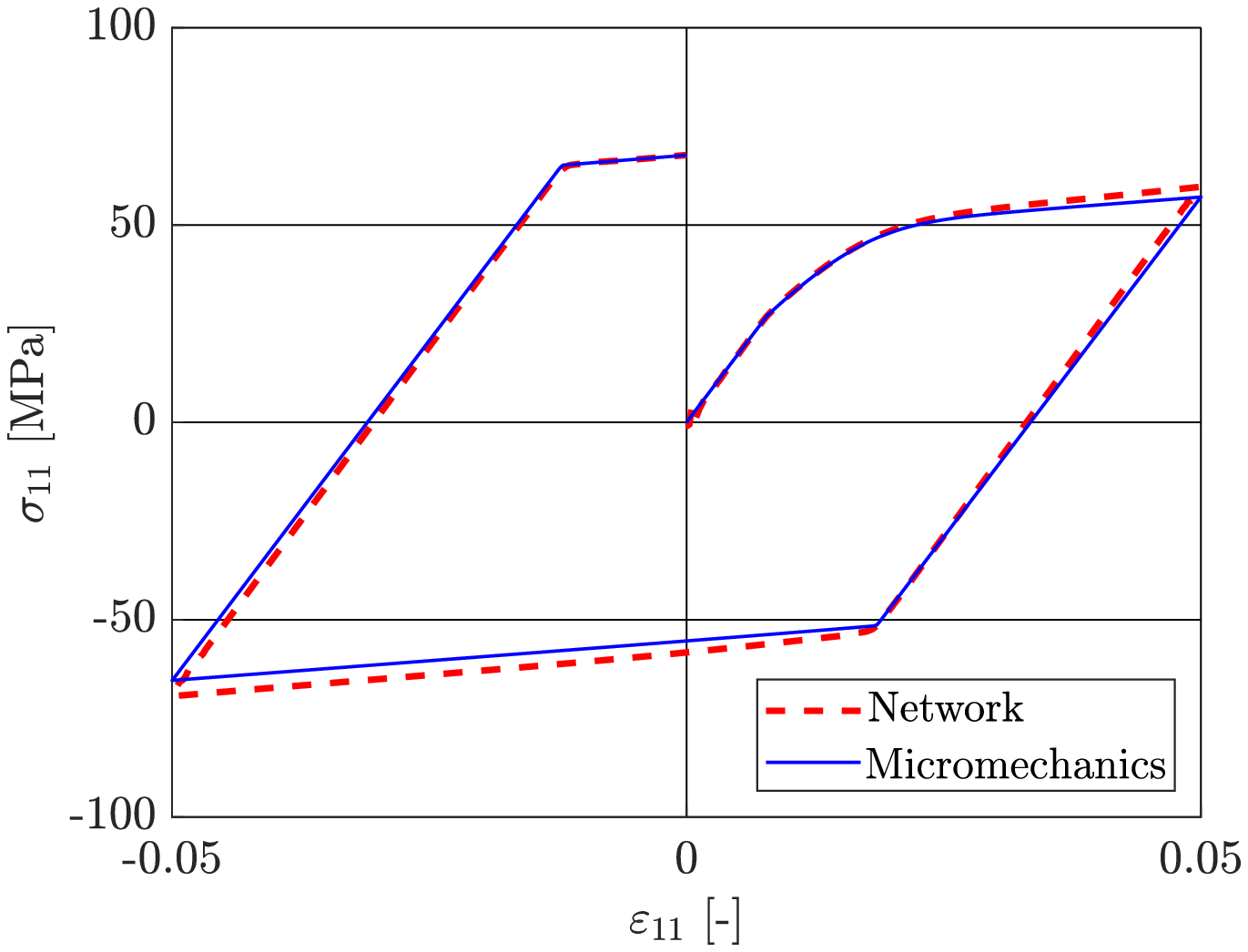}}
     \subfigure[]{\includegraphics[width=0.46\textwidth]{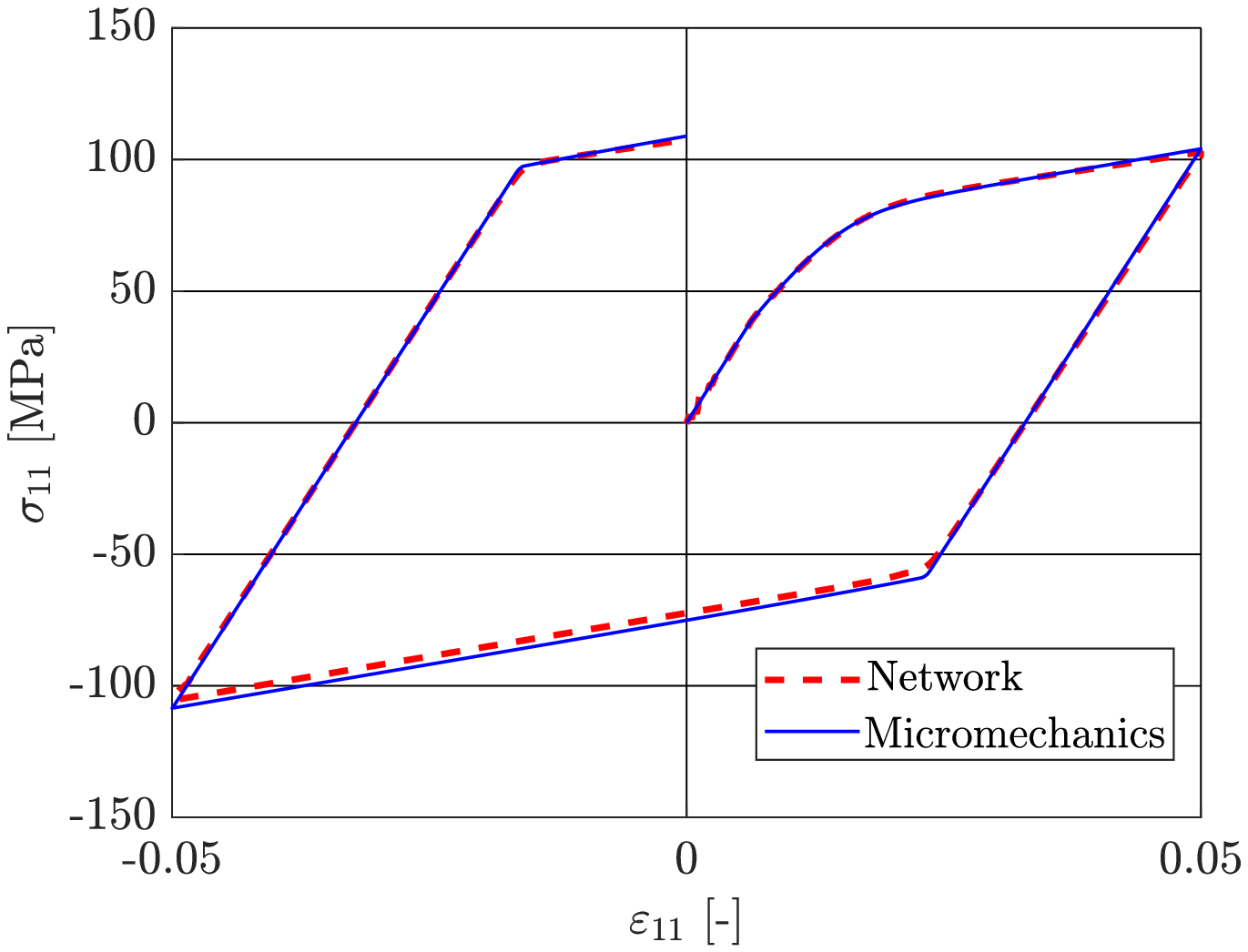}}
        \caption{Comparison of stress-strain curves obtained by the network and in micromechanical simulations, (a): a uniaxial test performed on a sample with a fiber volume fraction of 2.5\%, (b): a uniaxial test performed on a sample with fiber volume fraction of 20\%.}
        \label{fig:extrapolateStressStrain}
\end{figure}

The computed errors in the prediction of the $\sigma_{11}$ component, for different fiber volume fractions and different maximum control strains, are shown in Figure \ref{fig:extrapolateError}.
It seems like for fiber volume fractions in the permissible range (10\%-15\%), strains between 5\% and 7.5\% do not
lead to a very significant increase in the error. However, strains between 7.5\% and 10\% lead to a larger increase in the error for admissible fiber volume fractions. It can also be seen that the model generalizes very well to fiber volume fractions lower than 10\% and between 15\% and 20\% for strains in the admissible range ($\leq$5\%). The impressive extrapolation ability of the model for the admissible strains can be seen in the plot of two representative stress-strain curves in Figure \ref{fig:extrapolateStressStrain}. 
On the other hand, the detrimental effects on the prediction error for maximum strains larger than 5\% are amplified for fiber volume fractions outside the admissible range.

\section{Discussion}
\label{Discussion}
The acquired neural network, albeit exhibiting good performance, can not without a doubt be said to be optimal. In order to achieve an optimal network architecture, the hyper-parameters would need to be investigated more systematically. The complexity of the problem however results in very long network training times. A detailed survey of the hyper-parameters would require more dedicated hardware to enable parallel training of networks with multiple sets of parameters. 
It is possible to argue for the number of layers and neurons being appropriate. Previous attempts at predicting elasto-plasticity with GRU’s showed that networks with more than 3 layers and 500 neurons per layer did not display any significant improvement in performance \citep{Mozaffar2019}.

The chosen method of generating training data may be deemed successful. The created network shows good performance, and it seems that it is able to accurately predict the response to a wide variety of mechanical loads. There is however room for improvement. One obvious thing to investigate is the inclusion of characteristic training data that represent physically relevant phenomena such as hydrostatic loading. Alternatively, in future investigations it would be of interest to consider more smooth strain paths for training. This has been successfully implemented by others, using Gaussian processes \citep{Logarzo2021}.

A natural question to ask is what added value the proposed neural network model offers compared to the micromechanical model that it was trained on. To begin with, the proposed model offers the ability to change fiber orientation, and fiber volume fraction without having to perform additional simulations to homogenize the composite for the new set of parameters. This is important in the context of injection molding where the fiber density and orientation vary highly throughout the molded part. Secondly, the neural network model offers computational speed. As an example, using \textsc{digimat-mf} to compute the response to complex general 3D-loading (like the ones in the training data) may take up to a minute. In contrast, the network model makes the prediction in less than a second. Finally, the network contains all the information on loading/unloading and accumulated plastic strain/hardening purely through the strain history. Therefore, the process of determining these quantities repeatedly is not necessary. As a consequence, a lot of time is saved if the model is used in a framework where iterative calculations are required. This is important when implementing the network as a constitutive model in an FEM setting.

In the present work, an ANN model has been trained on micromechanical  simulations with the hope of the model being able to infer the physics of elasto-plasticity purely by observing stress-strain relations. As an alternative, a network may be trained with physically motivated constraints in order to ensure laws such as energy conservation. Linka et al. \citep{Linka2020} introduced Constitutive Artificial Neural Networks (CANN). These networks first compute the variants of the strain tensor and combine them with micromechanical  descriptors to compute a set of generalized invariants. The generalized invariants are in turn used to compute a strain energy functional. The stress and constitutive tensor are then easily obtained from the strain energy through differentiation. By computing the stress from an energy functional it is ensured that certain physical laws remain unviolated. The network is trained as usual with a strain input, with the goal of matching a predetermined stress-strain relation. One perk with that approach, in addition to certain laws of physics already being built into the network structure, is the reduced need for training data. 
As the model proposed in the present work yields some physically dubious results, such as accumulating plastic strain during hydrostatic loading, it would be of interest to investigate the possibility of introducing mechanically motivated constraints into future extensions of the network.

\section{Conclusions}
\label{Conclusions}
In this study, a deep neural network model that predicts the elasto-plastic response of a short fiber reinforced composites was developed. The material consists of elastic fibers with a $J_2$ elastoplastic matrix obeying a linear exponential hardening law. The network was trained on data from micromechanical simulations utilizing the Mori-Tanaka mean-field method for homogenization. The strain-paths used as input to the simulations were generated by utilizing a random walk in a 6D-strain space with bias directions. The created model allows for an arbitrary fiber distribution by defining a second order fiber orientation tensor. Additionally, it is possible to vary the fiber volume fraction between 10\% and 15\%, but it is possible to go slightly outside these bounds without introducing significant error.

The proposed model is computationally efficient and shows promising results. The model makes accurate predictions for a wide range of fiber orientation distributions and loading types. 
The model displays accurate yielding behavior with the appropriate hardening law (including saturation), and shows a clear difference in loading/unloading. It was furthermore demonstrated that the model correctly possesses rate-independence for a large range of loading rates. 
The developed model is a groundwork for general 3D micromechanics-based modeling of complex path-dependent behavior of SFRCs. In the next steps, this model will be extended by supplementary data for other matrices and fibers, and eventually by RVE finite element simulations for the sake of higher accuracy. Moreover, the possibility of improving the modeling by utilizing physical constraints in the model seems very promising. As an additional future outlook, a finite element implementation of the model is of interest. 



\section*{Acknowledgment}
S.M. Mirkhalaf is thankful for the financial support from the Swedish Research Council (VR grant: 2019-04715) and the University of Gothenburg. M.
Fagerstr\"om gratefully acknowledges the financial support through Vinnova’s
strategic innovation programme LIGHTer. Also, the authors would like to thank e-Xstream for providing a license of software \textsc{Digimat}.


\bibliography{bibliography}



\end{document}